%% file: ms.tex
\documentclass[runningheads]{llncs}
\usepackage[T1]{fontenc}
\usepackage[utf8]{inputenc}
\usepackage{graphicx}
\usepackage{algorithm}
\usepackage{algpseudocode}
\usepackage{biblatex}
\usepackage{amsmath}
\usepackage{float}
\begin{document}

\title{Efficient search of active inference policy spaces using k-means}

\author{Alex B. Kiefer\inst{1,2} \and
Mahault Albarracin\inst{1,3}}

\authorrunning{Kiefer and Albarracin}
\institute{VERSES Research Labs \and Monash University \and Université du Québec à Montréal }

\maketitle
\begin{abstract}
We develop an approach to policy selection in active inference that allows us to efficiently search large policy spaces by mapping each policy to its embedding in a vector space. We sample the expected free energy of representative points in the space, then perform a more thorough policy search around the most promising point in this initial sample.

We consider various approaches to creating the policy embedding space, and propose using $k$-means clustering to select representative points. We apply our technique to a goal-oriented graph-traversal problem, for which naive policy selection is intractable for even moderately large graphs.

\keywords{Active inference  \and Policy selection \and Hierarchical search}
\end{abstract}

\section{Introduction}

Active inference enjoys widespread popularity as a model for cognitive processes involving discrete decision-making. Typical implementations treat the process of active inference as a Partially Observed Markov Decision Processes (POMDP) \cite{smith2022step, 10.1162/NECO_a_00912, DBLP:journals/corr/abs-2201-03904}. This kind of model is subject to important limitations of scale, however. In particular, the time complexity of the exhaustive policy search carried out in standard POMDP active inference, in which the expected free energy (EFE) of each policy is computed out to a specified time horizon, renders it impractical for large state spaces involving many policies \cite{lueckmann2021benchmarking}.

There have been numerous efforts to address this limitation \cite{CHAMPION2022295}, including the exploration of tree search methods \cite{Fountasetal2020, whiteley2010efficient} and various methods of policy pruning \cite{da2020active}. Our contribution in this paper combines pruning with the use of vector space embeddings \cite{riesen2009graph} to create a structured policy representation in which qualitatively similar policies are proximal to one another. This representation can be exploited to conduct a fast search over representative points in the space, followed by a more thorough search in the neighborhood of the most promising candidate, yielding a hierarchical scheme for policy search related to ideas in hierarchical reinforcement learning \cite{DBLP:journals/corr/abs-2011-06335}.

The remainder of this paper is structured as follows: first, we briefly review the standard representation and selection of policies in POMDP active inference. We then give an overview of vector space embeddings, and consider how embedding strategies similar to those used in the domain of natural language may be applied to policies. We then consider how representative points in the space can be selected. Finally, we discuss experimental results in which we apply our technique to an active inference graph-traversal problem. In this domain, we show that it is possible to achieve accuracy comparable to exhaustive policy search with drastically lower time complexity.

\section{Policy selection in active inference}

This paper presupposes familiarity with the active inference framework, but we will briefly review the essentials of policy evaluation and selection in typical implementations. As is standard in other sorts of MDP models, policies in POMDP active inference are selected based on (a) the likelihood of states in the environment being realized, contingent on various actions (decisions) through which the agent can exert partial (probabilistic) control, and (b) the value to the agent of those states according to some value function. In a partially observed process, the effects of actions on states are not directly observed but are rather inferred from the states of observation channels representing sensory input \cite{smith2022step}.

The key difference between active inference and other POMDP models and in particular reinforcement learning models lies in the function used to compute the value of the policies \cite{millidge2020relationship}. In active inference, the standard objective (though see \cite{ParrFristonGeneralised, https://doi.org/10.48550/arxiv.1606.02396}) is to minimize expected free energy (EFE), which is the accumulation of the variational free energy of the system along future trajectories, given beliefs about the current environmental state plus a temporally deep generative model of how states are likely to evolve.

The expected free energy $G$ for a policy $\pi_i$ can be computed as
\begin{equation}
    G_{\pi_i} = \Sigma_{t\in{T}}\Big[{{D_{KL}[Q(o_t|\pi_i)||P(o_t)] + H(P(o_t|s_t))\dot{Q(s_t|\pi_i)}}}\Big]
\end{equation}
where $T$ is the time horizon, which is a hyperparameter of the model, $D_{KL}$ is a Kullback-Leibler divergence, $Q(o_t|\pi)$ and $P(o_t)$ are the expected approximate posterior and prior generative distribution, respectively, over observations at $t$, $H(P(o_t|s_t))$ is the entropy of the distribution over observations given states, and finally $Q(s_t|\pi)$ is the variational (approximate posterior) distribution over states \cite{schwartenbeck2013exploration}. Active inference differs from reinforcement learning in that minimizing EFE maximizes both an intrinsic reward (as defined by the generative density over observations $P(o)$ encoded the agent's "preference matrix") and information gain (the entropy term) \cite{Costa2020TheRB}.

To decide what to do, an active inference agent first infers the current state of the world from its observations (perceptual inference) \cite{schwartenbeck2013exploration}, then uses the inferred distribution over states to project the effects of action into the future, given a transition matrix that defines $p(s_{t+1}|s_t, u)$, where $u$ is a control state corresponding to a particular action. The distribution over observations at future time $t+1$ can then be calculated based on likelihoods $p(o|s)$. These observation probabilities are used to compute the EFE per policy, which in turn is used to select an action (see Appendix A for equations describing these updates).

\section{Structuring policy spaces with embeddings}

The serial calculation of the expected free energy over policies constitutes a serious bottleneck that renders even relatively small-scale models intractable given limited computational resources \cite{champion2022branching}. While performance can be improved by parallel processing, algorithmic efficiency is always welcome to complement raw compute power. In this paper we discuss a way to drastically increase the efficiency of policy search by applying the concept of a vector space embedding, in widespread use across machine learning, to policies for POMDP models.

An important caveat before proceeding is that our technique constructs a policy space from an initial enumeration of possible policies. This enumeration can itself represent a computational bottleneck which the present work does not aim to address. Moreover, the construction of embeddings can be computationally expensive, introducing a new bottleneck for large problems (see below). However, this overhead cost only needs to be computed once, rather than for every inference,  analogously to the training cost for a neural network, rather than during every iteration of a simulation, and scales with the existing policy-enumeration bottleneck.

\subsection{Vector space embeddings for policies}

In the most general terms, an embedding is a mapping from some set of items in a domain to points in a continuous vector space, with the important property that geometric and arithmetic relations among points in the space (such as Euclidean distance) capture corresponding relations among the mapped items \cite{liberti2014euclidean}.

Vector space embeddings were established as an essential tool for machine learning with the $word2vec$ model of \cite{DBLP:journals/corr/MikolovSCCD13}, which derived powerful vector representations for the domain of natural language processing via a simple local prediction task on large text corpora. Crucially, however, vector-space embeddings are a completely general modelling tool applicable in principle to any domain in which some regularity exists such that it can be exploited to construct the vector space. Image embeddings, for example, exploit the intrinsic structure of images (correlations among pixel values) from various domains \cite{https://doi.org/10.48550/arxiv.1707.05776}. In the case of discrete conventional symbol systems like natural language, the relevant structure exists in the corpus as extrinsic relationships among words and phrases \cite{li2018word}. 

Fundamentally, the problem of representing policies using vector spaces is similar to the case of language embeddings, if policies are thought of as sequences or more generally collections of actions. For example, consider the following three policies, defined over abstract actions $A$ - $G$:
\begin{align*}
    A \rightarrow B \rightarrow B \rightarrow C \\
    B \rightarrow A \rightarrow C \rightarrow C \\
    B \rightarrow E \rightarrow D \rightarrow G
\end{align*}
We may group these policies in several ways, based on the identities of the actions they contain. An approach analogous to the `bag-of-words' model in language processing, for example, would simply represent policies in terms of the counts of all possible actions that occur in them. By this metric, the top two policies above are more similar to one another than either is to the third, and would land closer together in the vector space.

Alternative embedding strategies might take into account the order in which actions occur. For example, a policy embedding might be constructed based on the occurrence or counts of $N$-grams (i.e. $A \rightarrow B$), or group together policies that begin or end with the same or similar actions. We propose and test a variation on the `bag of actions' approach, as well as an order-sensitive embedding strategy, in the experimental results section below.

One may also wish to use embeddings that explicitly incorporate expected value, e.g. by grouping together policies that allow the agent to achieve its goals (there is then a similarity to successor representations in reinforcement learning, which are applied to active inference in \cite{Millidge2022successor}). In order to examine what can be achieved on the basis of hierarchically organized policy spaces alone, without encoding additional information about rewarding states in the representation, we choose to focus instead on a purely structure-based grouping. This approach ought in principle to be very generally applicable, as it relies only on the assumption that, to some degree, structurally similar policies produce similar results.

\vspace{-0.2cm}
\section{Hierarchical policy selection}
The point of constructing an embedding, for present purposes, is to avoid having to compute the expected free energy of every possible policy. If we represent policies as points in a vector space, we can get a sense of the quality (from the agent's point of view) of the policies in various regions by sampling a small number of representative points, and computing their expected free energy. We can then begin from the most promising point (or top-$k$ points) and perform a more exhaustive local search. The challenge is then to ensure that we sample policies which are representative of the entire space while still reducing the amount of computation required.

\vspace{-0.2cm}
\subsection{Clustering with k-means}
In general, it is to be expected that the embeddings of real datapoints in a vector space will be packed into relatively dense clusters separated by gulfs corresponding to unlikely feature vectors. Visualizing the abstract vector spaces learned by neural networks using dimensionality reduction techniques such as t-SNE \cite{JMLR:v9:vandermaaten08a} often reveals precisely such separable clusters of datapoints.

Given the assumption of a non-uniform distribution of datapoints in the embedding space, algorithms such as $k$-means clustering \cite{goyal2014improving} may be effective in selecting a representative range of initial points to sample. $k$-means is a relatively simple unsupervised machine learning model closely related to the E-M algorithm  \cite{bottou1994convergence}, in which $k$ centroids are initialized (e.g. randomly) and each is assigned the datapoints closest to it according to some distance metric (e.g. Euclidean). The centroids are then re-calculated so as to minimize their mean distance to the datapoints to which they are assigned, and the assignment process repeats, converging on a solution in which the total distance between cluster centers (centroids) and datapoints is minimized. The optimized centroids are then effectively representative of different implicit `classes' of datapoints.

While one may consider many other clustering (and, more generally, unsupervised structure-discovery) algorithms, in the remainder of this paper we focus on $k$-means and demonstrate its effectiveness, in conjunction with a suitable policy embedding, on an applied active inference problem of some complexity.

\vspace{-0.2cm}
\subsection{Algorithms for policy selection}

Given the above, we propose two algorithms for hierarchical policy selection. The first begins by selecting a cluster of policies based on the EFE of its representative point (`cluster center'), then performs standard policy search within this cluster. The second instead samples $n$ points from each cluster from a uniform distribution, and a cluster is chosen for exploration based on the mean EFE of these points. Defining $\mathbf{E}$ as the embedding and $C_i$ and $c_i$ as policy cluster $i$ and representative point (i.e. the policy closest to the cluster centroid) of $C_i$, respectively, our basic algorithm is Algorithm \ref{alg1:cap} below. The alternative sampling-based algorithm is described in Appendix D.
\begin{algorithm}
\caption{Hierarchical policy selection}\label{alg1:cap}
\begin{algorithmic}
\State $(C, c)  \gets kmeans(\mathbf{E}, k)$ \Comment{output of the $k$-means algorithm}
\For{$0 <= $i < k}
\State $\mathbf{G}_{C_i} = EFE(c_i)$ \Comment{Standard EFE computation}
\EndFor
\State $\pi \gets \underset{C}{\operatorname{argmin}}(\mathbf{G}_{C})$
\State $u = select(\pi)$ \Comment{Standard active inference policy selection on reduced policy space}
\end{algorithmic}
\end{algorithm}

An interesting feature of this approach to hierarchical policy selection is that unlike other techniques that have been successfully applied to active inference such as MCTS \cite{Fountasetal2020}, it exploits purely structural features of the embedding space and does not require empirical tuning. That said, the quality of the solution depends heavily on choice of the hyperparameter $k$, as discussed in the following section. We note that many interesting variations on this idea remain to be explored, such as adding depth to the hierarchy (i.e. clusters of clusters) and sampling representative points from distributions informed by past performance.

\section{Experiment: Graph navigation}

We tested our proposal on a graph-navigation problem for which exhaustive policy search is impractical even on graphs of moderate complexity ($>6$ densely connected nodes). The agent's goal is to choose the shortest route to its desired destination on a random directed graph with weighted edges. While agent-based decision models are decidedly overkill for shortest-route discovery on graphs, this task provides an ideal environment in which to test our approach to policy search, since it requires inference within a temporally deep generative model and also offers a highly structured implicit policy space, as discussed below.

\subsection{Model}

We run our simulations on randomly generated directed, weighted graphs, with edge weights chosen from a small set of possible values. We first include each possible edge with probability $\frac{2}{|V|}$ (where $|V|$ is the number of nodes in the graph) and then enforce strong connectivity so that any node is reachable from any other. In addition, every node in the graph contains a self-connection. In each simulation, the agent is randomly assigned an initial location and `destination' node, and at each step chooses to move to an adjacent node or stay still. The self-connections have weights $>$ the largest between-node edge weight in the case of all but the `destination' node, whose self-connection weight is $0$. The agent has preferences for being at its destination and against traversing edges with large weights. To simplify the representation, we define the agent's possible locations in terms of edges on the graph, with the convention that the second node in an edge represents the agent's current location and the first node represents its previous location (full details of the agent's generative model are given in Appendix C). Because we were interested in modeling a situation in which agents knew how to reach their goals, we set the policy length to the number of nodes in the graph, so that at least one policy that reaches the goal is always available (given strong connectivity).

Since the states of the environment are encoded in terms of node pairs or edges, the actions that constitute our policies are transitions between edges. For example, $(A, B) \Rightarrow (B, C)$ is a valid policy on a graph with nodes $A$, $B$, and $C$ and directed edges $(A, B)$ and $(B, C)$. We prune policies containing `invalid' actions (that is, actions that imply impossible state transitions, such as moving directly from edge $(A, B)$ to edge $(C, D)$) from the policy space prior to search.

\subsection{Policy embeddings}

We experimented with three policy embedding strategies, an Edit Distance Matrix (EDM), a Bag-of-Edges (BOE) representation, and a BOE representation augmented with an extra dimension that records the terminal node of the policy (aBOE). 

The EDM for a list of policies is constructed by counting the number of moves it would take to transform one policy (i.e., path through a graph) into another. Intuitively, if the atomic moves are addition and deletion of nodes and edges, this should correspond to the number of elements (both nodes and edges) that appear in one graph and not the other. In other words, for edge sets $E_i$ and $E_j$ and vertex sets $V_i$ and $V_j$ of graphs $G_i$ and $G_j$, the edit distance $d_{ij}$ is:

    $$d_{ij} = |(V_i\cup{V_j} \setminus V_i\cap{V_j})| + |(E_i\cup{E_j} \setminus E_i\cap{E_j})|$$ 

\begin{figure}[H]
    \centering
    \vspace{-3cm}
    \includegraphics[width=\linewidth]{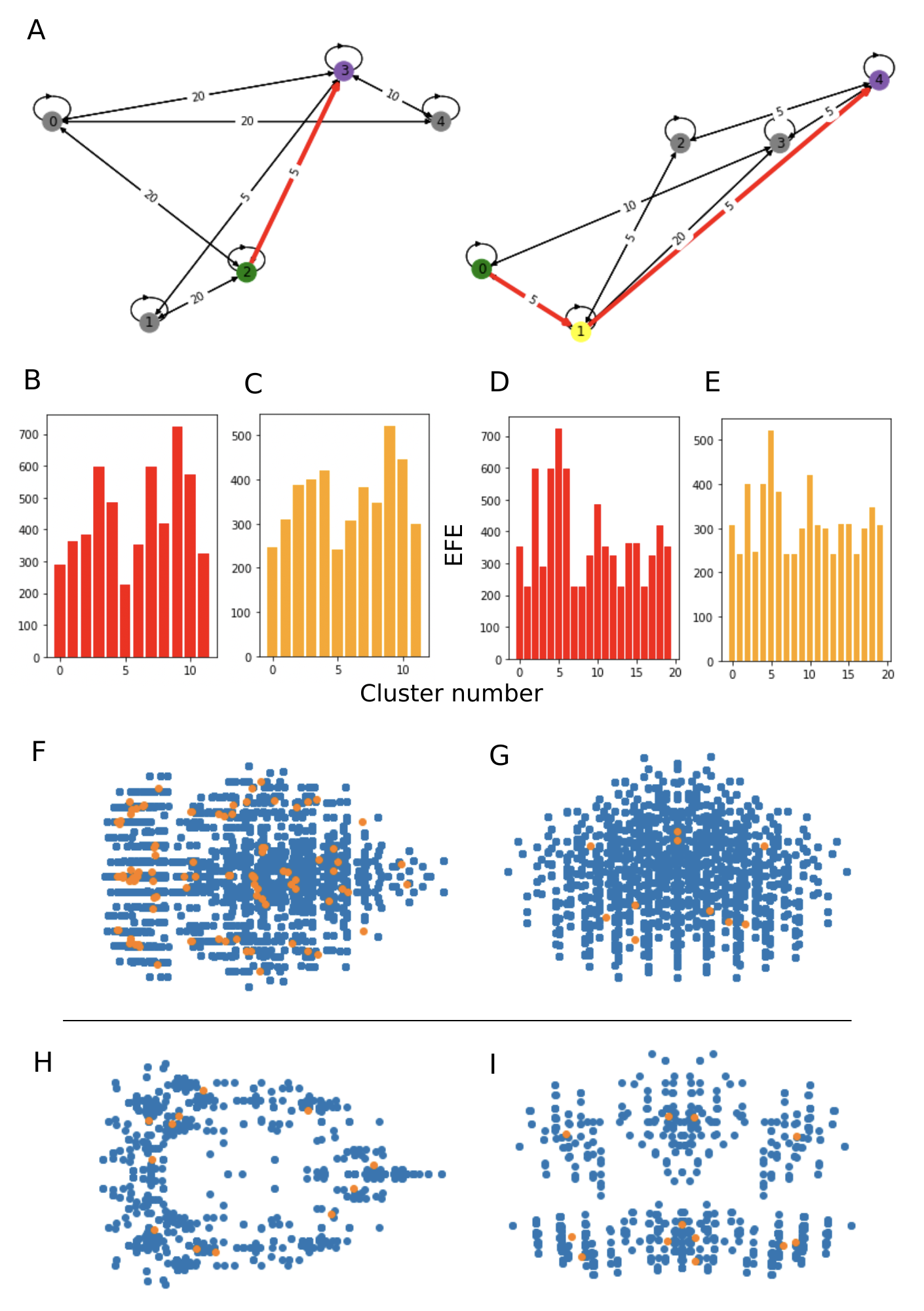}
    \caption{$\mathbf{A}$: Two sample graphs from our experiments, with start node (green) and destination node (purple) highlighted, along with the path the agent followed. Spatial layout is random and visual path distance does not track edge weight. $\mathbf{B}$: Expected free energy of policies represented by $k$ = 12 cluster centers. $\mathbf{C}$: Mean EFE of all clusters grouped by the corresponding cluster in plot ($\mathbf{B}$). $\mathbf{D}$: EFE of 20 randomly sampled cluster centers for $k$ = 100. $\mathbf{E}$: Mean EFE of corresponding clusters. $\mathbf{F}-\mathbf{I}$: 2D projections of policy embedding spaces using PCA (axes represent arbitrary dimensions in the reduced embedding space). Blue points are all policies, and orange points are cluster centers discovered by $k$-means. $\mathbf{F}$: Global edit distance matrix embedding. $\mathbf{G}$: Global `bag-of-edges' embedding. $\mathbf{H}$, $\mathbf{I}$: examples of corresponding local embeddings, limited to policies viable from a given location.}
    \label{data_image}
    \vspace{0cm}
\end{figure}

The embedding $\mathbf{d}_i$ for policy i is then the vector of edit distances to all other policies, which can combined with the other policy embeddings into a single embedding matrix $\mathbf{D}$.
 
The EDM representation is \emph{a priori} desirable because it takes the order of actions in a policy into account, but it can become computationally expensive to construct for larger graphs, leading to a different computational bottleneck to the one we set out to avoid. A much simpler representation is the Bag-of-Edges embedding, which as the name suggests is similar to the bag-of-words model in that it represents each policy simply by a vector of counts of the edges that occur in it. The augmented BOE embedding simply appends the identity of the last node reached by a policy to the BOE embedding for the policy.

\vspace{-0.4cm}
\subsection{Deriving representative points using $k$-means}

An implicit hypothesis behind our approach to policy search is that there would be a correlation between the degree to which two policies are structurally related (hence their location in embedding space) and their energy. To test this hypothesis, we plotted the EFE of each of the policies closest to the cluster centers, or of a random sample of them for large $k$, against the mean EFE of the policies in each cluster. We found that the degree to which the correlation holds is highly dependent on choice of $k$, but that with the right hyperparameter choice, the cluster centers are a good guide to the EFE in their regions (see Figure \ref{data_image}B-E).

Intuitively, assuming such correlation among the EFEs of policies proximal in the embedding space, there should be a tradeoff between the representativeness of the clusters chosen by $k$-means with respect to the EFE of their neighbors and efficiency gains due to using a small number of clusters for the initial sweep over policies. We found however that at least for the sizes of graph we explored (3-6 nodes), too large a $k$ value actually hurt performance as well as being less efficient. We performed a very limited hyperparameter search and found that a value of $k = 12$ worked well in practice, and report results for values $[6, 12]$.

As suggested by an anonymous reviewer, however, further optimization of $k$ is important, and a future avenue of research for this application would be to explore schemes for optimizing $k$ automatically, including online, e.g. so as to maximize EFE returns and minimize EFE variance within each cluster dynamically as the system evolves. This would bring our work more closely in line with \cite{Fountasetal2020}, in which Monto Carlo tree search and an amortized variational inference procedure are used to improve the efficiency of policy selection.

\vspace{-0.5cm}
\subsection{Local VS global embeddings}
We found that running $k$-means on the full embedding matrix for the entire policy space (including policies that were `invalid' from the agent's current location) sometimes returned no, or very few, valid policies among the cluster centers, leading to sub-optimal choices. 

To remedy this, we try pruning all policies not beginning at the agent's current location by defining local embeddings $\mathbf{E}_{s_i}$ for each possible agent location $s_i$ as $\mathbf{E}_{s_i} \gets \{\mathbf{e}_j \in \mathbf{E}: s_i \in \pi_j \}$ and perform clustering on these reduced subspaces to get per-location $k$-means clusters and cluster centers. At each simulation step we then run our hierarchical search algorithm on the appropriate subspace. For a fair comparison, we benchmarked this procedure against standard active inference policy selection run on the same local policy subspaces.

The utility of embeddings is best evaluated by measuring their performance, but for visualization purposes we also constructed dimensionally reduced representations of the high-dimensional embedding spaces using Principal Component Analysis (PCA), shown in Figure \ref{data_image}F-I. Though heuristic, these plots suggest that policies do indeed cluster in interesting ways, and that the $k$-means procedure is good at finding these clusters.

\vspace{-0.3cm}
\subsection{Results}

\begin{figure}[H]
    \centering
    \vspace{-0.2cm}
    \includegraphics[width=\linewidth]{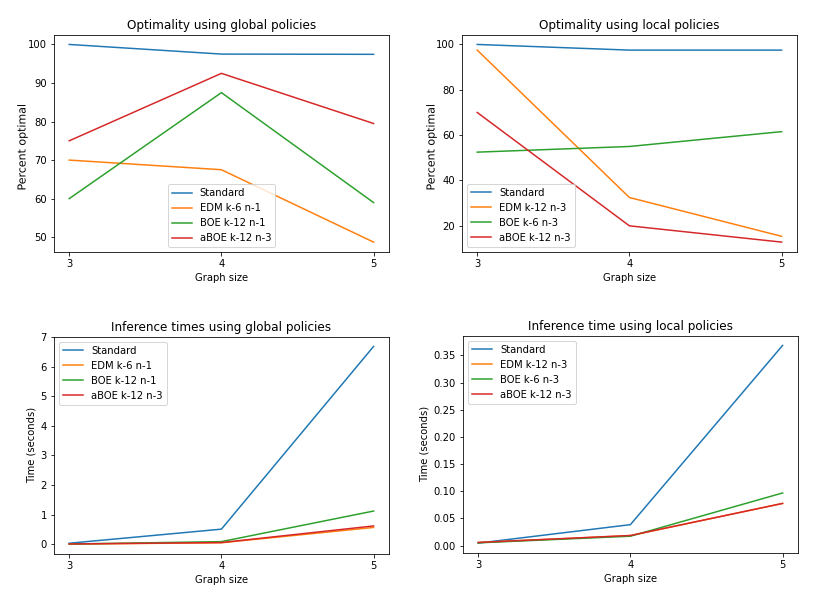}
    \vspace{-0.2cm}
    \caption{Selected results on graphs with $3-5$ nodes, for `Global' (full policy space) and `Local' (location-based subspace) conditions. Top row: Percent of solutions found that were optimal. $Standard$ is standard active inference policy search, and the hyperparameters used for the embedding conditions are listed next to the embedding name. Bottom row: Mean policy inference times for standard VS embedding conditions.}
    \label{results_image}
    \vspace{0cm}
\end{figure}

\vspace{-0.2cm}
A sample of our results is presented in Figure \ref{results_image}  (please see Appendix E for additional data). To obtain these results, we generated 40 random graphs in each size category ($3, 4, 5$) and computed mean execution time and optimality for each embedding type (including "None" / standard policy selection, EDM, BOE and aBOE), as well as for two values of the hyperparameters $k$ (number of clusters) and $n$ (number of samples used to calculate per-cluster EFE). An optimal solution was defined as one in which the agent takes a shortest path from its initial location to its `destination node' and then remains there. Figure \ref{results_image} shows only one hyperparameter combination for each embedding type.

For larger graphs, hierarchical policy search improved the calculation time for action selection by about an order of magnitude, with similar relative reductions in the global and local conditions. The mean inference time using k-means/embeddings was about .17 seconds, and for standard policy inference, 1.2 seconds. Construction times for all embeddings, including the time to carry out k-means clustering, were negligible for all embedding styles except EDM, which exponential time complexity precluded using on graphs of size > 5 nodes. The EDM representation achieved near-optimal results on small graphs, but its performance in any case degraded on larger graphs. While these preliminary results are not in general impressive in terms of optimality, the best results suggest that our technique could be made competitive with more extensive hyperparameter tuning (as well as potentially different clustering and embedding methods).

\begin{figure}[h]
    \centering
    \includegraphics[width=\linewidth]{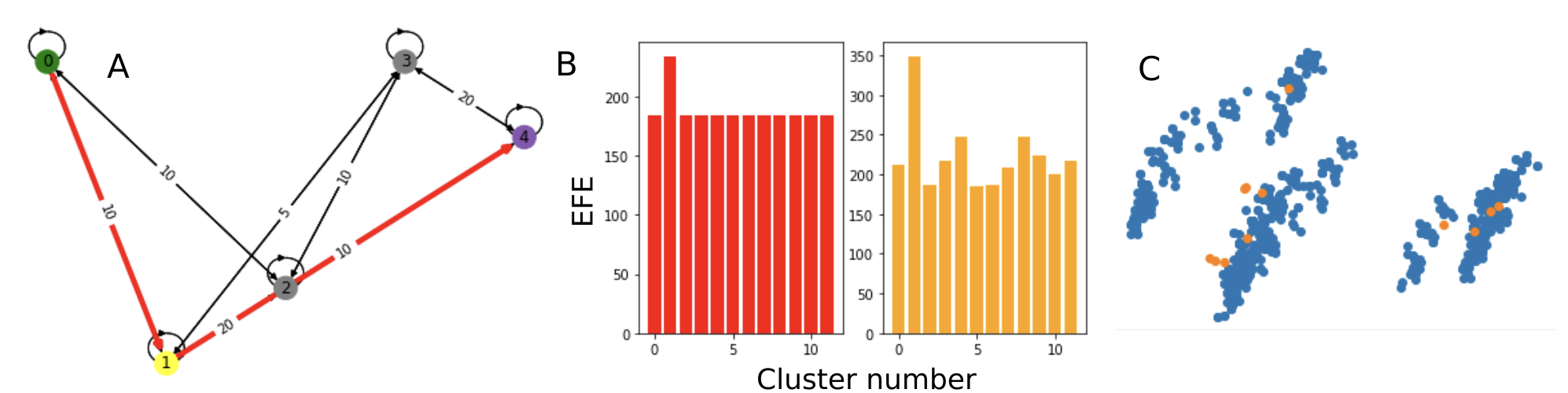}
    \caption{$\mathbf{A}$: Graph in which the agent failed to find a route. $\mathbf{B}$(Left, Right): Cluster center and mean cluster EFE values, respectively. $\mathbf{C}$: Policy embeddings and cluster centers for this example.}
    \label{failure}
    \vspace{0cm}
\end{figure}

\vspace{-0.4cm}
We analyzed one interesting failure case in which the agent did not move from its initial location. As shown in Fig. \ref{failure}, it appears that in this case, the EFE of the clusters was not a good guide to the local energy landscape, and in addition, $k$-means did not adequately cover the policy space (note the obvious cluster on the left without an assigned cluster center).  

\vspace{-0.4cm}
\section{Conclusion}

The experiments reported in this paper are very much an initial cursory exploration of the application of embedding spaces and clustering to hierarchical policy search in active inference. Very sophisticated graph embedding schemes, using neural networks trained on random walks for example, could be applied to problems like ours. The initial results we report suggest that this line of research may prove useful in expanding the sphere of applicability of active inference POMDP models. 

\section*{Acknowledgements}
Alex Kiefer and Mahault Albarracin are supported by VERSES Research.

\section*{Code Availability}

Code for reproducing our experiments and analysis can be found at: 
\\\text{https://github.com/exilefaker/policy-embeddings}

\printbibliography

\section*{Appendix A: Computing per-policy EFE}

The expected free energy for a policy can be computed as follows. First, we infer a distribution over states at the current step of the simulation, combining the transition probabilities and current observations:

$$Q(s_t) = \sigma{[\ln{A_{o_t}} + \ln{B_u\dot{Q(s_{t-1})}}]}$$

The inferred distribution over states can then be used to project the effects of action into the future, given a parameter $B$ that defines $p(s_{t+1}|s_t, u)$, where $u$ is a control state corresponding to a particular action:

$$Q(s_{t+1}) = B_u\dot{Q(s_t)}$$

Given an inferred distribution over future states, the distribution over observations at future time $t+1$ can be calculated as

$$Q(o_{t+1}) = A\dot{Q(s_{t+1})}$$
where $A$ is the likelihood mapping from (beliefs about) states to observations. The above assumes a single observation modality and controllable state factor, but can straightforwardly be generalized to larger factorized state spaces and multiple observation channels.

By repeating the above process using the $Q(s_{t+1})$ resulting from the previous time-step as input to the next, and accumulating the $G_{\pi_i}$ defined in Eq. (1) out to the policy horizon, we can derive a distribution $Q_{\pi}$ over policies as $\sigma({-\mathbf{G}_\pi})$, where $\mathbf{G}_\pi$ is the vector of expected free energies for all available policies and $\sigma(x)$ is a softmax function. Finally, the next action is sampled from a softmax distribution whose logits are the summed probabilities under $Q_{\pi}$ of the policies consistent with each action.

Note that in the above we have omitted aspects of typical active inference models not material to our concerns in this paper, such as precision-weighting of the expected free energy, habits, and inverse temperature parameters.

\section*{Appendix B: Policy selection and expected free energy}

We use the standard procedure outlined in the Introduction for selecting policies with one exception: the expected free energy has an additional term which is the dot product of the posterior distribution over states (locations) with the associated edge weights. We combine this with the standard EFE calculation using a free hyperparameter $\lambda$:

\begin{equation}
    G_{\pi_i} = \Sigma_{t\in{T}}\Big[{{D_{KL}[Q(o_t|\pi_i)||P(o_t)] + H(P(o_t|s_t))\dot{Q(s_t|\pi_i)}}}\Big]+ \lambda*{weights\cdot{Q(s_t|\pi)}}
\end{equation}

This choice was made purely for convenience since otherwise preferences over weights would have to be represented using an awkward categorical distribution, and it has no impact on the main comparison between policy search techniques of interest to us in this paper.

\section*{Appendix C: Generative model details} 

In our experimental setup, an active inference agent's generative model is automatically constructed when a graph is generated. The standard variables in active inference POMDPs have the following interpretations in our model:

\begin{itemize}
    \item states: An edge ($node_{prev}$, $node_{current}$) in the graph. We interpret the first node in the edge as the agent's previous location and the second node as its current location.
    \\
    \item observations: A tuple (($node_{prev}$, $node_{current}$), $weight$) representing observations of edges and corresponding edge weights, where the edge corresponds to the node pair in $states$
    \\
    \item control states: There is an action (and corresponding control state) for every possible local transition in the graph.
    \\
    \item A: The agent's `A' or likelihood matrix, which in this case is simply an identity mapping from states to observations
    \\
    \item B: The state transition matrix, which encodes deterministic knowledge of action-conditioned state transitions, and is constructed so as to exclude invalid transitions (i.e. between non-adjacent nodes in the graph)
    \\
    \item C: Preference matrix, which distributes probability mass equally over all edges that end on the agent's `destination' node. There is also implicitly a preference against high edge weights, but to simplify the representation we incorporate this directly within the expected free energy calculation (see Appendix B).
    \\
    \item D: Prior over initial location states.

\end{itemize} 

\section*{Appendix D: Sample-based hierarchical policy selection} 

With variables defined as above, but with $c_j$ denoting the $j$th randomly sampled policy in a cluster, the alternative sample-based policy selection algorithm is:

\begin{algorithm}[H]
\caption{Sample-based hierarchical policy selection}\label{alg2:cap}
\begin{algorithmic}
\For{$0 <= i < k$}
\For{$0 <= j < n$}
\State $c_j \sim U(C_i)$ \Comment{This is a uniform distribution over cluster members}
\EndFor
\State $\mathbf{G}_{C_i} = \frac{{\sum}_j{EFE(c_j)}}{n}$
\EndFor
\State $\pi \gets \underset{C}{\operatorname{argmin}}(\mathbf{G}_{C})$
\State $u = select(\pi)$
\end{algorithmic}
\end{algorithm}

\vspace{-0.5cm}
\section*{Appendix E: Additional results} 

\begin{figure}[H]
    \centering
    \includegraphics[width=\linewidth]{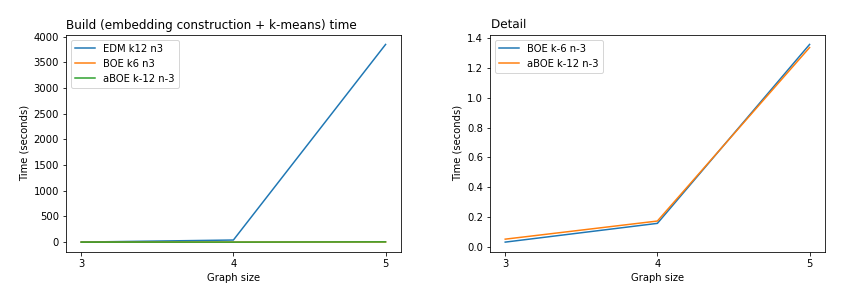}
    \caption{Left: Time taken to construct embedding spaces and perform k-means clustering on the resulting embeddings. The increased times for both construction and clustering for the EDM representation are due to the relatively much larger dimensionality of the EDM embedding: one dimension for each policy, rather than one for each vertex and edge, as in the BOE and aBOE representations. Right: `detail' plot of the construction times by graph size for the BOE and aBOE embeddings.}
    \label{construction_times}
    \vspace{0cm}
\end{figure}

Here we present some additional experimental results. Figure \ref{construction_times} plots the combined embedding construction and k-means clustering times for each embedding type. Table \ref{tab:1} below shows the full set of optimality results we obtained, averaged across trials (i.e. across particular graphs in each category). Here, "None" denotes standard policy selection. Best embedding results for each graph size are bolded.

\input{accuracy_data2.tex}

\end{document}

%% file: accuracy_data2.tex
\begin{table}
\centering
\caption{Percent of solutions optimal}
\label{tab:1}
\def\arraystretch{1.1}
\setlength\tabcolsep{5.0pt}
\begin{tabular}{llccrrr}
      Graph size     &      & & &     3 &    4 &    5 \\
     \hline \hline \\
Scope & Embedding & \textit{k} & \textit{n} &       &      &      \\
\hline \\
      \vspace{0.3cm}
Global & None & --- & --- & 100.0 & 97.5 & 97.4 \\
      & BOE & 6 & 1 &  70.0 & 65.0 & 56.4 \\
      &      &      & 3 &  77.5 & 62.5 & 46.1 \\
      &      & 12 & 1 &  60.0 & 87.5 & 59.0 \\
      \vspace{0.2cm}
      &      &      & 3 &  77.5 & 62.5 & 51.3 \\
      & EDM & 6 & 1 &  70.0 & 67.5 & 48.7 \\
      &      &      & 3 &  80.0 & 72.5 & 56.4 \\
      &      & 12 & 1 &  67.5 & 72.5 & 64.1 \\
      \vspace{0.2cm}
      &      &      & 3 &  80.0 & 72.5 & 61.5 \\
      & aBOE & 6 & 1 &  82.5 & 85.0 & 66.7 \\
      &      &      & 3 &  87.5 & 85.0 & 61.5 \\
      &      & 12 & 1 &  85.0 & 67.5 & 35.9 \\
      \vspace{0.7cm}
      &      &      & 3 &  75.0 & \textbf{92.5} & \textbf{79.5} \\
      \vspace{0.3cm}
Local & None & --- & --- & 100.0 & 97.5 & 97.5 \\
      & BOE & 6 & 1 &  17.5 & 37.5 & 48.7 \\
      &      &      & 3 &  52.5 & 55.0 & 61.5 \\
      &      & 12 & 1 &  42.5 & 47.5 & 41.0 \\
      \vspace{0.2cm}
      &      &      & 3 &  82.5 & 42.5 & 25.6 \\
      & EDM & 6 & 1 &   5.0 & 52.5 & 45.0 \\
      &      &      & 3 &  50.0 & 65.0 & 35.0 \\
      &      & 12 & 1 &  50.0 & 60.0 & 40.0 \\
      \vspace{0.2cm}
      &      &      & 3 &  \textbf{97.5} & 32.5 & 15.4 \\
      & aBOE & 6 & 1 &  15.0 & 12.5 &  5.1 \\
      &      &      & 3 &  35.0 & 12.5 &  5.1 \\
      &      & 12 & 1 &   5.0 & 22.5 & 12.8 \\
      &      &      & 3 &  70.0 & 20.0 & 12.8 \\
\end{tabular}
\end{table}